%% file: workflow.tex
\documentclass[letterpaper]{article}
\usepackage{aaai}
\usepackage{times}
\usepackage{helvet}
\usepackage{courier}
\usepackage{graphicx}
\usepackage{graphicx}
\usepackage[table,xcdraw]{xcolor}
\usepackage{booktabs}
\usepackage{diagbox}

\newcounter{enumcounter}

\nocopyright

\begin{document}

\title{Workflow Complexity for Collaborative Interactions:\\ \emph{Where are the Metrics?} -- A Challenge}
\author{Kartik Talamadupula \and Biplav Srivastava \and Jeffrey O. Kephart\\
Human-Agent Collaboration Group\\
IBM T. J. Watson Research Center\\
Yorktown Heights, NY 10598\\
{\tt \small \{krtalamad, biplavs, kephart\} @ us.ibm.com}
}
\maketitle


\input{abstract}


\input{introduction}

\input{related}
\input{examples}

\input{background}
\input{discussion}

\input{conclusion}

\small
\bibliographystyle{named}
\bibliography{workflow}

\end{document}

%% file: abstract.tex
\begin{abstract}
\begin{quote}


In this paper, we introduce the problem of denoting and deriving the complexity of workflows (plans, schedules) in collaborative, planner-assisted settings where humans and agents are trying to jointly solve a task. The interactions -- and hence the workflows that connect the human and the agents -- may differ according to the domain and the kind of agents. We adapt insights from prior work in human-agent teaming and workflow analysis to suggest metrics for workflow complexity. The main motivation behind this work is to highlight metrics for human comprehensibility of plans and schedules. The planning community has seen its fair share of work on the synthesis of plans that take diversity into account -- what value do such plans hold if their generation is not guided at least in part by metrics that reflect the ease of engaging with and using those plans?

\end{quote}
\end{abstract}

%% file: introduction.tex
\section{Introduction}
\label{sec:introduction}



The emergence of the Internet and  application (app) centric service-oriented platforms for various kinds of consumer tasks have resulted in an explosion in the interactions between humans and automated agents that assist them in tasks. Given their large number, a formal measurement of the inherent complexity of these interactions is desirable to assist in the design of useful and efficient decision-making algorithms and systems.

\begin{figure*}
  \centering
	\includegraphics[width=0.8\textwidth,height=9cm]{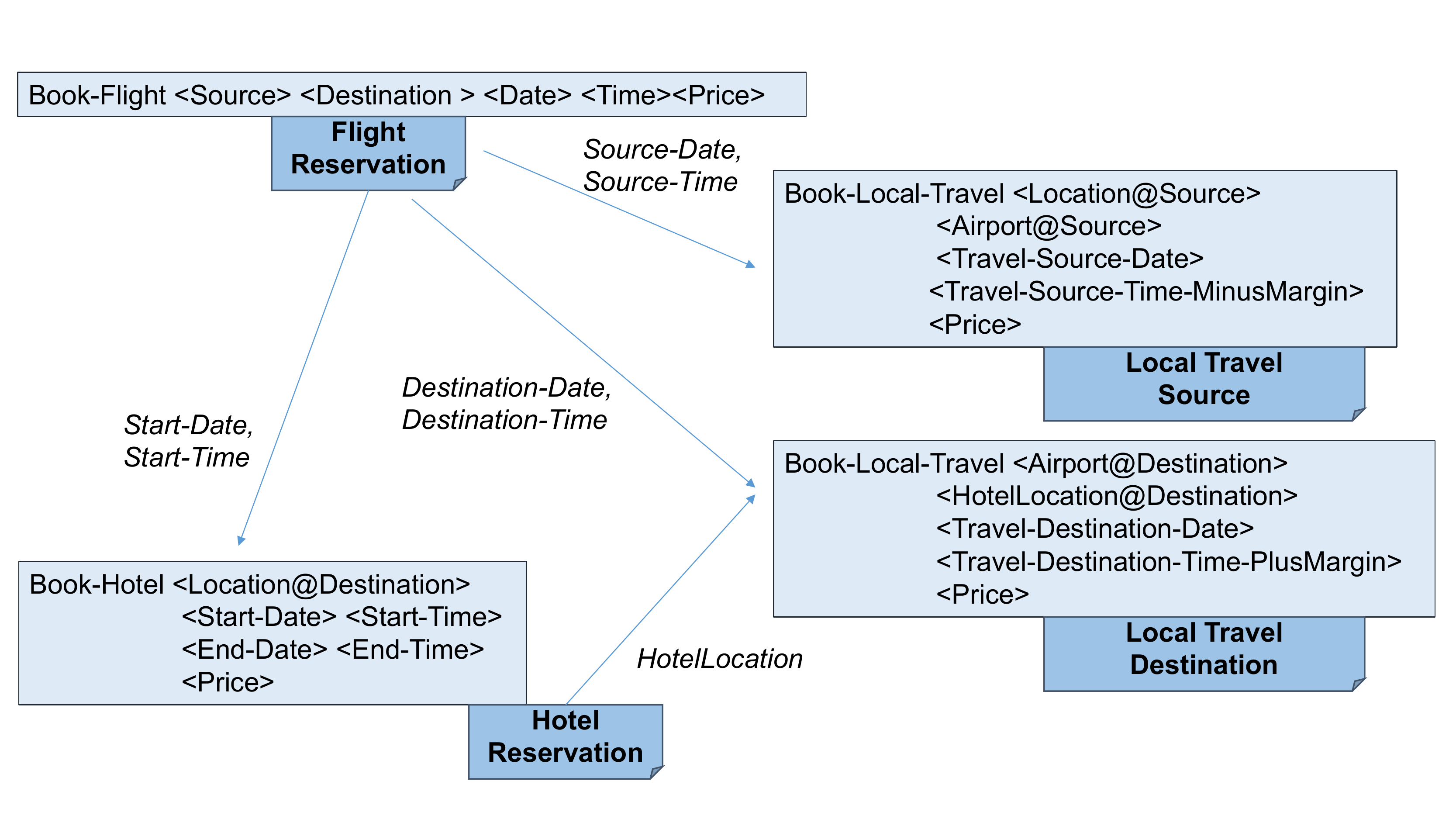}
  \caption{Actions, data (parameters) and constraint variables in a small travel example.}
  \label{fig:eg-sketch}
\end{figure*}

We present a usecase that illustrates the kinds of interactions we are discussing -- consider a person living in New York who wants to book a travel itinerary for a short personal trip to Seattle. The workflow for planning this trip will consist of a flight reservation, hotel reservation, and reservation for local travel in the source and destination cities. These bookings could each be made via websites, over a dialog interface, via IoT interfaces,
or manually over the phone with a travel agency. Each communication modality introduces its own constraints and complexity.  We highlight the workflow complexity in this specific example using Figure~\ref{fig:eg-sketch}. Here, three action instances are shown for booking a flight, a hotel, and local travel. The data artifacts are the booking confirmations, whose variables constrain the other actions in the workflow. In the workflow fragment that is shown, local travel at the destination is most constrained as it depends on the flight's arrival time, as well as the location of the hotel at the destination.
In general, the flight booking will result in dates (and times) which create a dependency for the hotel reservation. Finally, flight and hotel reservations give the date and locations for which local travel needs to be booked. The overall complexity of booking this short leisure trip may differ from a business trip, where meeting schedules have to be taken into account; and may further differ from an international trip where the processing of travel documents has to be taken into account.


In such scenarios, automation faces two main challenges. The first is the problem of knowledge acquisition and engineering pertaining to the domain of interest -- in the travel scenario above, this knowledge would constitute the various actions available to the agent to create a successful workflow, and the dependencies between those actions. This kind of problem is the purview of the flourishing Knowledge Engineering for Planning \& Scheduling (KEPS) community. The second major problem is that of explaining the plan and the interactions underlying it to the (human) user/consumer of the plan. An important sub-problem in this is \emph{measuring} the complexity of the said interaction -- without such measures, an automated system that is trying to aid in such interactions will be unable to distinguish between and rank workflows of vastly differing complexities that all achieve the same goal. Complexity measures provide the ability to rank the planner's mediation in such scenarios, and allow the planner to produce directed help that will enable easier achievement of the user's goals. We highlight this second problem in this paper.

%% file: related.tex
\section{Prior Work}
 
There is a rich body of work on workflow representation,
composition, and execution~\cite{workflow-book}. Over the past decade, there
have been approaches for semi or fully automated composition of workflows using planning that look at control and data driven issues~\cite{Srivastava03webservice}. However, much of this work is in the context
of single agent decision-making. There is no 
prior work, to our knowledge, that characterizes the 
complexity of workflows in a collaborative setting. 

The planning community has also seen advances in the problems of measuring the distance between plans~\cite{roberts2014evaluating,goldman2015measuring}, and generating diverse plan alternatives~\cite{nguyen2012generating}. However, very little research has focused on exactly what the different metrics that go into creating diverse plans should be. Such work has mostly looked at measures (cost, duration, robustness, etc.) that treat the plan as an artifact disconnected from humans, who must execute, understand, or participate in that workflow. Humans typically perceive complexity both from interaction issues, as well as from the actions in a workflow. Indeed, there is a long history of prior work from a linguistic and structural perspective for the former~\cite{dialog-complexity-ux}. However, there has been no focus on creating a class of metrics that attempt to define the complexity of a plan or workflow. We intend this paper as a challenge to the community to do exactly that. 

%% file: examples.tex
\section{Workflow Complexity: Example Usecases}
\label{sec:examples}

We described the Travel Booking usecase in Section~\ref{sec:introduction}; here, we describe some other collaborative examples to highlight complexities that an automated decision making system can help reduce.

\subsubsection{Scheduling Meetings}

A common collaborative task in the workplace is deciding a meeting time and venue, given a topic. This mundane task is complicated by the fact that there are different roles for participants in the meeting, hard and soft scheduling constraints, and limited access to participant information which changes with context. In such scenarios, setting up a meeting between colleagues who are at the same level organizationally may be more complex than one convened by the head of the organization -- in the former there may be more hard constraints and various alternatives  have to be considered, while in the latter everyone is likely to mark their (conflicting) constraints as soft. An automated agent~\cite{cranshaw2017calendar} can play a crucial role in improving the efficiency of meeting scheduling\footnote{This is distinct from the actual scheduling problem, which is to find a satisfying assignment given everyone's constraints -- our problem considers the workflow of scheduling the meeting.}. Specifically, it can verify participants and roles, identify potential conflicts from existing schedules, ask (the fewest number of) people to re-visit their constraints, and explain alternative time-slots.

\subsubsection{Evaluating Hiring Choices}

Another workplace example is the evaluation of a set of candidates by a multi-disciplinary panel of experts. The experts may evaluate the candidate's technical skills, non-technical (soft) skills, organizational fit, HR concerns, career progression etc. Depending on the role, the process may involve many interview rounds, evaluations, and discussion. Further complexity is added by variations in the evaluation scales, disagreements among the experts, and relative weights of selection criteria. An automated agent can make this process more efficient by formalizing the contributions of the experts, focusing the team on key decision factors, retrieving relevant candidate data, eliminating human bias, and providing justifications to the stakeholders when asked. 





\begin{table*}[htp]
\scriptsize
\centering
\resizebox{\textwidth}{!}{%
\begin{tabular}{lcccccccc}
\rowcolor[HTML]{CCCCCC} 
{\color[HTML]{FFFFFF} \textbf{\diagbox{\color[HTML]{000000}USECASE}{\color[HTML]{000000}METRIC}}} & {\color[HTML]{000000} \textbf{NT}} & {\color[HTML]{000000} \textbf{IT}} & {\color[HTML]{000000} \textbf{AD}} & {\color[HTML]{000000} \textbf{FO}} & {\color[HTML]{000000} \textbf{Com}} & {\color[HTML]{000000} \textbf{EC}} & \textbf{PC} & \textbf{MC} \\
\rowcolor[HTML]{EFEFEF} 
{Travel Booking} & {\color[HTML]{000000} \textbf{L}} & {\color[HTML]{000000} \textbf{H}} & {\color[HTML]{000000} \textbf{H}} & {\color[HTML]{000000} \textbf{H}} & {\color[HTML]{000000} \textbf{H}} & {\color[HTML]{000000} \textbf{H}} & {\color[HTML]{000000} \textbf{M}} & {\color[HTML]{000000} \textbf{M}} \\
\cellcolor[HTML]{FFFFFF}{Scheduling Meetings} & \cellcolor[HTML]{FFFFFF}{\color[HTML]{000000} \textbf{H}} & \cellcolor[HTML]{FFFFFF}{\color[HTML]{000000} \textbf{M}} & \cellcolor[HTML]{FFFFFF}{\color[HTML]{000000} \textbf{M}} & \cellcolor[HTML]{FFFFFF}{\color[HTML]{000000} \textbf{L}} & \cellcolor[HTML]{FFFFFF}{\color[HTML]{000000} \textbf{H}} & \cellcolor[HTML]{FFFFFF}{\color[HTML]{000000} \textbf{L}} & {\color[HTML]{000000} \textbf{H}} & {\color[HTML]{000000} \textbf{H}} \\ 
\rowcolor[HTML]{EFEFEF} 
{Evaluating Hiring Choices} & {\color[HTML]{000000} \textbf{L}} & {\color[HTML]{000000} \textbf{H}} & {\color[HTML]{000000} \textbf{H}} & {\color[HTML]{000000} \textbf{L}} & {\color[HTML]{000000} \textbf{H}} & {\color[HTML]{000000} \textbf{M}} & {\color[HTML]{000000} \textbf{H}} & {\color[HTML]{000000} \textbf{H}} \\ 
\cellcolor[HTML]{FFFFFF}{Human-Robot Teaming} & \cellcolor[HTML]{FFFFFF}{\color[HTML]{000000} \textbf{M}} & \cellcolor[HTML]{FFFFFF}{\color[HTML]{000000} \textbf{H}} & \cellcolor[HTML]{FFFFFF}{\color[HTML]{000000} \textbf{M}} & \cellcolor[HTML]{FFFFFF}{\color[HTML]{000000} \textbf{M}} & \cellcolor[HTML]{FFFFFF}{\color[HTML]{000000} \textbf{L}} & \cellcolor[HTML]{FFFFFF}{\color[HTML]{000000} \textbf{M}} & {\color[HTML]{000000} \textbf{L}} & {\color[HTML]{000000} \textbf{L}} \\  
\rowcolor[HTML]{EFEFEF} 
{Medical Treatment} & {\color[HTML]{000000} \textbf{H}} & {\color[HTML]{000000} \textbf{L}} & {\color[HTML]{000000} \textbf{L}} & {\color[HTML]{000000} \textbf{L}} & {\color[HTML]{000000} \textbf{H}} & {\color[HTML]{000000} \textbf{M}} & {\color[HTML]{000000} \textbf{H}} & {\color[HTML]{000000} \textbf{H}} \\ 
\cellcolor[HTML]{FFFFFF}{Personal Finance} & {\color[HTML]{000000} \textbf{M}} & {\color[HTML]{000000} \textbf{M}} & {\color[HTML]{000000} \textbf{H}} & {\color[HTML]{000000} \textbf{L}} & {\color[HTML]{000000} \textbf{H}} & {\color[HTML]{000000} \textbf{H}} & {\color[HTML]{000000} \textbf{H}} & {\color[HTML]{000000} \textbf{H}} \\ 

\end{tabular}%
}
\caption{Workflow complexity metrics and their footprint; H - High, M - Medium, L - Low; Metrics described in Section~\ref{sec:metrics}.}
\label{metrics-table}
\normalsize
\end{table*}

\subsubsection{Human-Robot Teaming}

Planning for human-robot teaming (HRT)~\cite{talamadupula2010planning,chakraborti2016planning} considers the problem of humans and robots in goal-oriented environments, and the planner's mediation through control of the robotic agent. HRT scenarios usually involve extensive interaction between the human and the robot. Automated mediation can make the teaming more efficient in a number of ways, including coordination to reduce communication~\cite{talamadupula2014coordination}, and restricting the number of agents that a human has to deal with.

\subsubsection{Deciding a Medical Treatment Plan}

Another illustrative collaborative task, from the area of health, is deciding a medical plan for a person given a health condition (initial state) and a desirable new condition (goal state). For example, if a pregnant person has to be operated on for a planned child birth, specialists of the concerned medical fields 
need to coordinate specific procedures; schedule it with relevant nursing staff; complete insurance formalities; and reserve resources like the operation room.  Some of these processes follow standardized or regulated workflows, while others are case-specific depending on patient risk factors, etc. Furthermore, the data in such scenarios must be controlled due to confidentiality and regulatory reasons~\cite{leyens2017use}. An automated decision maker can help by focusing the attention of the medical team on ensuring compliance, examining risk factors and medical requirements, and avoiding costly mistakes that may foreclose future remedial actions.



\subsubsection{Personal Finance}

Increasingly, personal finance has emerged as an area of great opportunity as well as challenge for decision making systems and decision assistants. Usecases like buying a house, saving for retirement, or filing one's taxes are important decisions with long-term life implications. A number of characteristics must be considered including the various alternatives available, their costs (both immediate and future), legal and compliance issues, etc. A specific example of such a decision making scenario is an automated tax assistant -- such an assistant must be aware of the tax code which prescribes various rules and regulations that must be followed, and must recommend the best tax plan while optimizing a number of metrics including minimizing amount paid as tax, minimizing the complexity of the plan, and maximizing compliance (to minimize the chances of audits and fines).




%% file: background.tex
\section{Metrics}
\label{sec:metrics}

We now list some metrics from prior work that can be adapted to the problem we consider. Chakraborti et al. (\citeyear{chakraborti2016formal}) provide a framework for studying and evaluating interaction between human and robot team-members in goal-oriented environments. Some useful metrics that can be adapted from that work are: 


\begin{enumerate}
\item \textbf{Neglect Tolerance (NT)}: How long the agent is able to perform well without human intervention.
\item \textbf{Interaction Time (IT)}: Time spent in communication.
\item \textbf{(Robot) Attention Demand (AD)}: Measures the attention demanded by the agent.
\item \textbf{Fan Out (FO)}: Communication load on the humans; proportional to the number of agents.
\item \textbf{Compliance (Com)}: How well the actions of an agent convey its intention to comply.
\setcounter{enumcounter}{\value{enumi}}
\end{enumerate}



\noindent Separately, Keller et al. (\citeyear{config-complexity}) consider
the problem of workflow complexity relating to configuring Information Technology (IT) infrastructure, e.g. a web application. They define configuration complexity as ``the complexity of carrying out a configuration procedure as perceived by a human system manager''; and track information along three dimensions, which are (respectively) analogous to control flow, data flow, and space complexity in software engineering: 


\begin{enumerate}

\setcounter{enumi}{\value{enumcounter}}
\item \textbf{Execution Complexity (EC)}: Number of actions and context switches.
\item \textbf{Parameter Complexity (PC)}: Number of parameters used by actions, and their usage variations.
\item \textbf{Memory Complexity (MC)}: Number of configuration values which need to be remembered along the workflow, and over the actions.

\end{enumerate}

\noindent These measures are all relevant from a human-agent collaboration perspective, as they relate to the effort needed to review a plan and to gain human trust.

%% file: discussion.tex
\section{Discussion}



In Table~\ref{metrics-table}, we present the above metrics juxtaposed with their footprint in the collaborative domains introduced in Section~\ref{sec:examples}. The footprint itself is quantized into three categories -- \textbf{High (H)}, \textbf{Medium (M)}, and \textbf{Low (L)}.  We address a number of points in relation to the table. First and foremost, the table should be read column-wise, for each metric. Second, the High/Medium/Low annotations denote the typical or average-case profile for that metric in the respective usecase, and may vary depending on the specific problem instance etc. 

Third, these values do not represent any intrinsic \emph{goodness} -- high neglect tolerance is good in scenarios like Human-Robot Teaming, because it shows that the automated agent is more independent; while low compliance might be a bad thing if the human wants constant confirmation or reassurance from the agent, like in medical treatment scenarios. However, these can easily switch depending on the domains and users in question: medical professionals may want a less independent agent (lower neglect tolerance), while meeting scheduling agents may not be required to show all the steps of their work. A general rule-of-thumb is that if the metric profile of a particular usecase is reflected in the plans that a planner produces, overall team success is more likely.


We now discuss the metrics from Table~\ref{metrics-table} in the context of creating new metrics that define the complexity of plans or workflows in terms of the interaction issues, as well as the complexity of the actions that constitute those workflows. The first set of metrics informally represent interaction issues: Neglect Tolerance (NT), Interaction Time (IT), and Attention Demand (AD) are related to each other, and are concerned with the demands that a workflow imposes on the user/human via the agent's roles in the workflow. Similarly, IT and Fan Out (FO) offer a measure of the communication that is expected from the user, and how many different agents the user must accommodate (the assumption being that communication load increases as a function of the number of such agents). The second set of metrics represents the complexity of the actions in the workflow itself: while a scenario that involves scheduling meetings might feature a number of possible alternative workflows and each action might consist of multiple parameters, other scenarios like human-robot teaming might in fact feature relatively fewer alternatives and action parameters. These are all important to track in the final plan that is generated for the human-agent team, since they contribute to the difficulty of explaining the workflow and its constituent parts (as required).

%% file: conclusion.tex
\section{Conclusion \& Future Work}

We conclude by reiterating that the metrics we discuss in this paper differ from the traditional metrics used in the planning community, which apply specifically to actions and goal-states; the optimal profiles for these metrics are instead at least partially  determined by the usecase in question. We would like to use these as a starting point in ultimately creating metrics that explain the complexity of the workflow cumulatively from the perspective of the agent that must understand, explain, or execute it. Our hope is that this paper will spur action in two directions: (1) the post-processing of plans from existing planners to take cumulative plan complexity metrics into account; and eventually, (2) the creation of new planners that can handle such complexity metrics directly in the state-space search and plan synthesis processes.